\documentclass[runningheads]{llncs}
\usepackage[T1]{fontenc}
\usepackage{graphicx}
\usepackage{amssymb}
\usepackage{amsmath}
\usepackage{hyperref}

\usepackage{array}
\usepackage{booktabs}
\usepackage{colortbl} 
\usepackage{xcolor}   
\usepackage{tabularx} 
\usepackage{makecell} 
\usepackage{pifont} 
\newcommand{\cmark}{\ding{51}} 

\begin{document}
\titlerunning{DS MYOLO: A Reliable Object Detector}
\title{DS MYOLO: A Reliable Object Detector Based on SSMs for Driving Scenarios}
\author{Yang Li \and
	Jianli Xiao*}
\authorrunning{Li and Xiao}
\institute{University of Shanghai for Science and Technology, Shanghai, China, 200093 \\ 
	Email: \texttt{223330831@st.usst.edu.cn} (Yang Li); \texttt{audyxiao@sjtu.edu.cn} (Jianli Xiao)}
\maketitle              % typeset the header of the contribution
\begin{abstract}
Accurate real-time object detection enhances the safety of advanced driver-assistance systems, making it an essential component in driving scenarios. With the rapid development of deep learning technology, CNN-based YOLO real-time object detectors have gained significant attention. However, the local focus of CNNs results in performance bottlenecks. To further enhance detector performance, researchers have introduced Transformer-based self-attention mechanisms to leverage global receptive fields, but their quadratic complexity incurs substantial computational costs. Recently, Mamba, with its linear complexity, has made significant progress through global selective scanning. Inspired by Mamba's outstanding performance, we propose a novel object detector: DS MYOLO. This detector captures global feature information through a simplified selective scanning fusion block (SimVSS Block) and effectively integrates the network's deep features. Additionally, we introduce an efficient channel attention convolution (ECAConv) that enhances cross-channel feature interaction while maintaining low computational complexity. Extensive experiments on the CCTSDB 2021 and VLD-45 driving scenarios datasets demonstrate that DS MYOLO exhibits significant potential and competitive advantage among similarly scaled YOLO series real-time object detectors.

\keywords{Driving Scenarios \and Object Detection \and SSM \and YOLO.}
\end{abstract}
\section{Introduction}
%\subsection{A Subsection Sample}
In recent years, the rapid development of deep learning has continuously injected new energy into the field of object detection. In autonomous driving scenarios, real-time detection and accurate identification of traffic signs and vehicle identities are crucial for enhancing the safety of driving systems\cite{cheng2023towards}. However, in driving scenarios, targets often vary significantly in scale and size, leading to poor visual features and susceptibility to noise interference. This makes object detection one of the most challenging tasks in autonomous driving. CNNs, with their parameter sharing and optimized hardware acceleration, have made significant progress in real-time object detectors. However, their local focus makes it difficult to effectively capture targets of different scales in driving scenarios, limiting their performance. Therefore, developing a high-performance real-time object detector is an important and meaningful endeavor.

In the past, general object detection paradigms primarily focused on CNN-based two-stage detection networks, such as Faster R-CNN\cite{ren2015faster}, Mask R-CNN\cite{he2017mask}, and Cascade R-CNN\cite{cai2018cascade}. However, the pre-generation of candidate region proposals in two-stage detectors often results in inadequate real-time performance. Recently, research in object detection has increasingly shifted towards end-to-end single-stage detection algorithms, such as YOLO\cite{redmon2016you}, SSD\cite{liu2016ssd}, CornerNet\cite{law2018cornernet}, and FCOS\cite{tian2020fcos}. Single-stage detection models feature simpler architectures, with the YOLO series models, in particular, achieving a commendable balance between speed and accuracy. This has garnered significant attention from both the academic and industrial communities.

The YOLO networks, especially from YOLOv3\cite{redmon2018yolov3} onwards, typically consist of three main structures: backbone, neck, and head. The backbone extracts deep features from input images. For instance, YOLOv3, YOLOX\cite{ge2021yolox}, YOLOv7\cite{wang2023yolov7}, and YOLOv8\cite{ultralytics2023} use Darknet-53\cite{redmon2018yolov3}, while YOLOv4\cite{bochkovskiy2020yolov4} and YOLOv5\cite{yolov5v6.1} use CSPDarknet-53\cite{bochkovskiy2020yolov4}. YOLOv6\cite{li2022yolov6} employs EfficientRep\cite{li2022yolov6}, and YOLOv9\cite{wang2024yolov9} uses the lightweight GELAN. The neck structure fuses multi-scale features to enhance multi-scale representation capabilities. SPPELAN\cite{wang2024yolov9} optimizes multi-scale feature extraction efficiency, and PAN\cite{he2015spatial} enhances feature fusion based on FPN\cite{lin2017feature}. The head structure decodes the features from the neck to generate final detection results, evolving from anchor-based (e.g., YOLOv5\cite{yolov5v6.1}, YOLOv7\cite{wang2023yolov7}) to more efficient anchor-free (e.g., YOLOv6\cite{li2022yolov6}, YOLOv8\cite{ultralytics2023}, YOLOv9\cite{wang2024yolov9}) and NMS-free (YOLOv10\cite{wang2024yolov10}) designs.

Object detectors based on the Transformer encoder-decoder architecture, such as the DETR\cite{carion2020end} series, leverage the global feature modeling capabilities of the self-attention mechanism to achieve performance comparable to state-of-the-art detectors. However, the quadratic computational complexity poses challenges in balancing speed and accuracy. Inspired by the effectiveness of attention mechanisms, channel attention mechanisms based on CNNs, such as SE\cite{hu2018squeeze}, ECA\cite{wang2020eca}, and their variants\cite{fu2019dual}\cite{hou2021coordinate}, have also demonstrated significant gains. Recent research has shown that methods based on State Space Models (SMMs), such as Mamba\cite{gu2023mamba}\cite{dao2024transformers}, have achieved remarkable success in visual tasks due to their powerful global modeling capabilities and linear complexity advantages\cite{zhu2024vision}\cite{wang2024mamba}\cite{pei2024efficientvmamba}.

Inspired by previous works, we propose a novel object detector named DS MYOLO. This detector integrates a Simplified Volitional Scan Fusion Block (SimVSS Block) to achieve deep global feature fusion, and introduces an Efficient Convolutional Operator (ECAConv) to address the shortcomings of the Standard Convolution(SC) in cross-channel interactions. We validate the superiority of DS MYOLO on the publicly available CCTSDB 2021\cite{zhang2022cctsdb} traffic sign dataset and the VLD-45\cite{yang2021vld} vehicle logo dataset. Experimental results demonstrate that DS MYOLO exhibits strong competitiveness among state-of-the-art detectors of similar scale. In summary, our contributions can be outlined as follows:

1)	To further enhance detection performance through feature fusion, we design a Simplified Volitional Scan Fusion Block (SimVSS Block) to achieve deep global feature fusion. This block consists of a State Space Model (SMM) in series with a feedforward network, enhanced by residual connections, effectively integrating global and local features.

2)	We propose an Efficient Channel Attention Convolutional Operator (ECAConv). By decoupling the channels post-convolution and performing cross-channel attention interactions, ECAConv significantly establishes dependencies between channels and enhances representation, while maintaining computational complexity similar to SC.

3)	We further design different scales of DS MYOLO (-N/-S/-M) real-time object detectors based on the proposed SimVSS Block and ECAConv. On the CCTSDB 2021\cite{zhang2022cctsdb} and VLD-45\cite{yang2021vld} traffic scene datasets, DS MYOLO demonstrates robust competitiveness compared to existing state-of-the-art real-time object detectors.

\section{Related works}
\subsection{Real-time Object Detectors}
With the rapid development of autonomous driving, developing real-time and efficient object detectors is crucial for real-world applications. To balance speed and accuracy, researchers have dedicated significant time and effort to developing efficient object detectors. Among these, the YOLO series models have garnered widespread attention due to their simple structure and end-to-end detection characteristics. Starting from the initial YOLOv3\cite{redmon2018yolov3}, the architectural design of backbone-neck-head networks has been a key factor in enhancing model performance. YOLOv4\cite{bochkovskiy2020yolov4}], based on CSPNet\cite{wang2020cspnet}, optimized the previously used DarkNet backbone structure\cite{redmon2018yolov3} and introduced a series of data augmentation methods\cite{bochkovskiy2020yolov4}\cite{yun2019cutmix}. YOLOv5\cite{yolov5v6.1} incorporated strategies such as adaptive anchor box computation and automated learning rate adjustment. YOLO-X\cite{ge2021yolox} employed a label assignment strategy (SimOTA) and introduced a decoupled head to further improve training efficiency and detection performance. YOLOv6\cite{li2022yolov6} integrated re-parameterization methods into the YOLO architecture to balance accuracy and speed. YOLOv7\cite{wang2023yolov7} introduced the Extended Efficient Layer Aggregation Network (E-ELAN) as the backbone to further enhance performance. YOLOv8\cite{ultralytics2023} focused on analyzing the shortcomings of previous YOLO models and achieved higher performance by integrating their strengths. Gold-YOLO\cite{wang2024gold} proposed the GD mechanism to improve multi-scale object fusion performance. YOLOv9\cite{wang2024yolov9} introduced the GELAN backbone and enhanced the model's expressive capabilities through PGI. YOLOv10\cite{wang2024yolov10} proposed a dual-label assignment strategy without NMS, improving the overall efficiency of the model.
\subsection{Transformer-base object detection}
Transformers\cite{vaswani2017attention}, with their self-attention mechanism, excel in addressing long-range dependency issues. DETR\cite{carion2020end} was the first to apply the Transformer architecture to object detection, simplifying the pipeline by eliminating manually designed anchor boxes and NMS components, garnering significant attention. However, DETR's training convergence remains inefficient.  Subsequently, Deformable-DETR\cite{zhu2020deformable} improved upon DETR by combining deformable convolutions with self-attention calculations, effectively accelerating convergence. Conditional DETR\cite{meng2021conditional} introduced the Conditional Cross-Attention mechanism to expedite DETR's training. DAB-DETR\cite{liu2022dab} utilized dynamic anchor boxes directly as queries in the Transformer decoder, enhancing training speed and inference performance. Anchor DETR\cite{wang2022anchor} incorporated anchor-based query design and Row-Column Decoupled Attention (RCDA), achieving comparable performance to DETR while improving efficiency. DN-DETR\cite{li2022dn} introduced a query-denoising training method to accelerate DETR's training process and further enhance performance. Group DETR\cite{chen2023group} employed a group-based training strategy with one-to-many assignments to increase training efficiency. RT-DETR\cite{zhao2024detrs} proposed an efficient hybrid encoder architecture by separating intra-scale interactions and cross-scale fusion, further improving model efficiency and accuracy. Rank-DETR\cite{pu2024rank} introduced a rank-oriented architecture design, significantly boosting inference precision.
\subsection{SSMs-Based Vision State Space Model}
Recently, Mamba\cite{gu2023mamba}\cite{dao2024transformers} has garnered significant attention for its linear complexity in addressing long-range dependency problems. Subsequently, Vision Mamba\cite{zhu2024vision} was the first to apply the SSM to visual backbone networks, achieving performance comparable to, or even surpassing, Vision Transformers (ViT). VMamba\cite{liu2024vmamba} introduced the Cross-Scan Module (CSM) to capture the global receptive field, enhancing visual representation with linear computational complexity. LocalMamba\cite{huang2024localmamba} proposed a local scanning strategy to strengthen feature dependencies within local windows while maintaining a global perspective. EfficientVMamba\cite{pei2024efficientvmamba} combined efficient selective scanning with convolution in the backbone, achieving a balance between accuracy and efficiency. MambaOut\cite{yu2024mambaout} explored the necessity of SSM in visual tasks, experimentally validating SSM's higher value for tasks with long sequences and autoregressive characteristics, and providing foundational support for downstream tasks like segmentation. MSVMamba\cite{shi2024multi} introduced a multi-scale scanning mechanism, enhancing the ability to learn dependencies across different resolutions. Inspired by Mamba's outstanding contributions to various visual tasks, we integrated the SSM module into our network's feature fusion, achieving significant performance enhancement.
\section{Method}
\begin{figure}
	%\vspace{-2em} %\vspace{-2em} % Remove vertical space before the image
	\includegraphics[width=\textwidth]{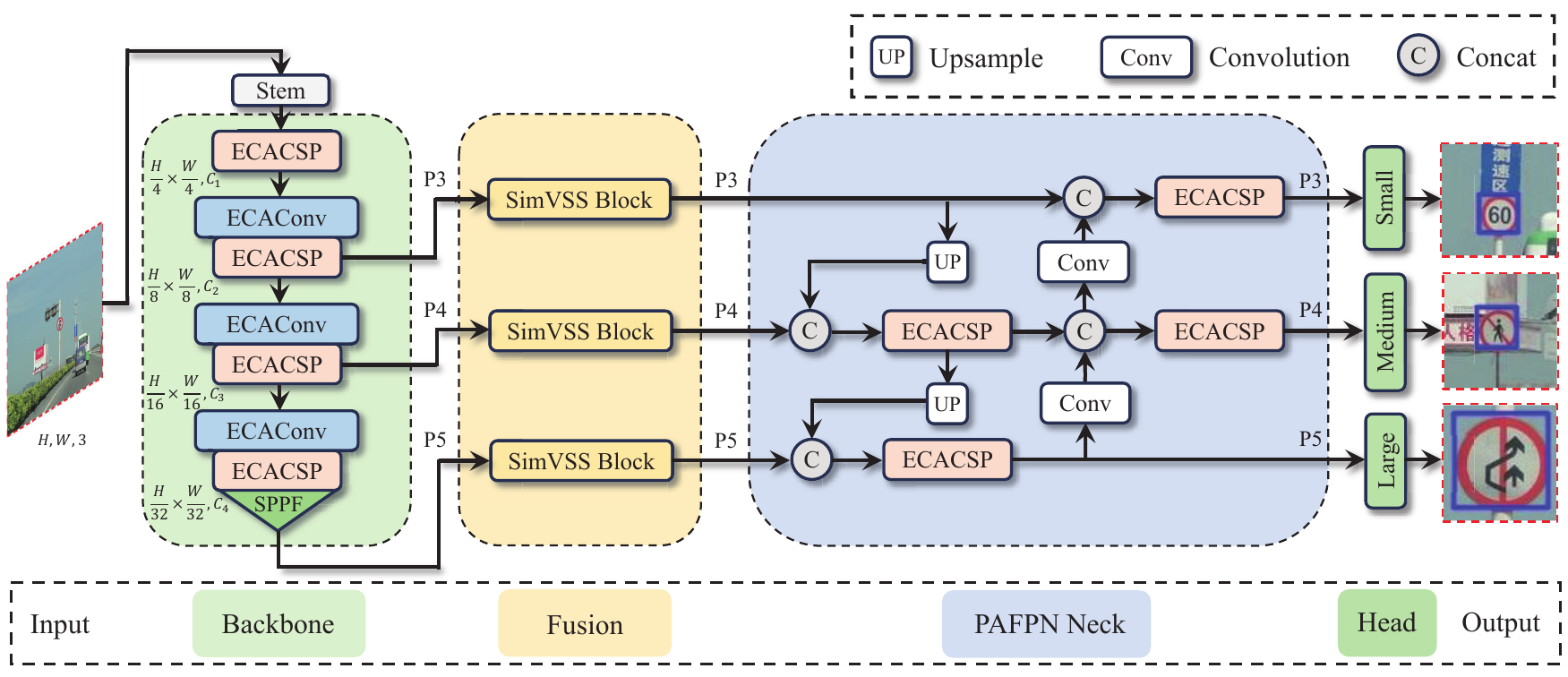}
	\caption{Overall architecture of DS MYOLO.} \label{fig1}
	\vspace{-2em} % Remove vertical space before the image
\end{figure}
\subsection{Overall Architecture of DS MYOLO}
The overall architecture of DS MYOLO is illustrated in Figure\ref{fig1}. In the backbone network, the Stem is composed of SC, batch normalization, and a SiLU activation function, stacked sequentially and downsampled twice, resulting in a 2D feature map with dimensions $(\frac{H}{4},\frac{W}{4})$, and ${C_i}$ channels. To effectively extract rich features in the backbone network, ECAConv is used for downsampling with a stride of 2, and ECACSP is employed to further extract abundant local features. Our object detection model introduces a fusion layer before the neck network. This fusion layer uses three SimVSS Blocks to achieve deep integration of feature layers $\left\{ {P3,P4,P5} \right\}$ while maintaining low computational complexity. In the neck, we follow the PAFPN\cite{ultralytics2023} approach, using $3 \times 3$ SC for downsampling with a stride of 2 and further integrating local features through ECACSP. We adopt a practical decoupled head and NMS-free design\cite{wang2024yolov10}, which effectively decodes small, medium, and large targets in the input, enabling efficient detection across different scales.
\subsection{Fusion Layer Based on SimVSS Block}
The traditional YOLO model transmits features extracted by the backbone network directly to the neck network for feature communication. While this method effectively enhances the salience of local features, it overlooks the feature dependencies within the global receptive field. Previous research has demonstrated that increasing the receptive field can beneficially enhance model performance. Given the larger feature map size of shallow networks, we employ a simplified SimVSS Block based on SSM to process the output features of the backbone network. The fused global features are then subjected to nonlinear transformations through a forward network to improve the model's fitting capacity.
\begin{figure}
	%\vspace{-2em} %\vspace{-2em} % Remove vertical space before the image
	\includegraphics[width=\textwidth]{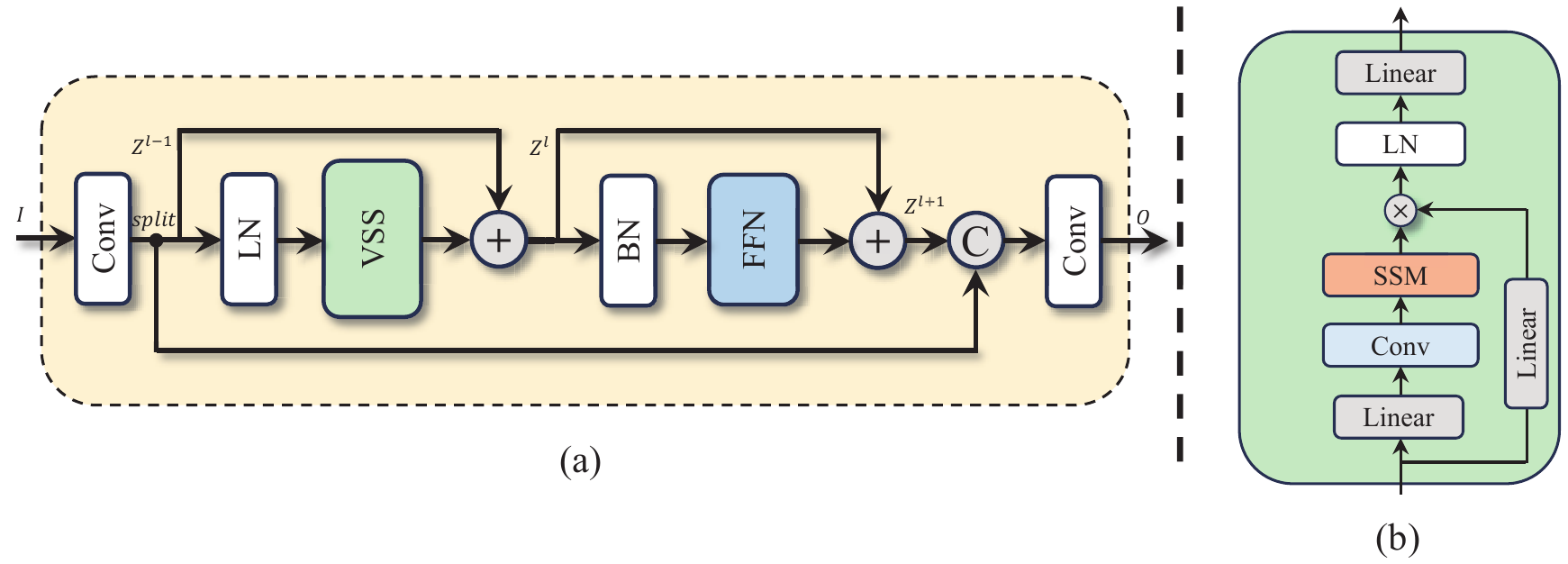}
	\caption{Detailed structure of SimVSS Block ((a) represents component modules of SimVSS Block, (b) represents key internal architecture of VSS module).} \label{fig2}
	\vspace{-2em} % Remove vertical space before the image
\end{figure}

The structure of SimVSS Block is illustrated in Fig.\ref{fig2}. The primary design is based on the SSM and a feedforward network, with residual connections and normalization layers included to stabilize gradient training and accelerate model convergence. A traditional SSM can be viewed as a linear time-invariant system function that maps a univariate sequence $x(t) \in \mathbb{R}$ to an output sequence $y(t) \in \mathbb{R}$ via an intermediate hidden state $h(t) \in {\mathbb{R}^N}$. Given the state transition matrix $A \in {\mathbb{R}^{N \times N}}$ as the evolution factor, the weight matrix and the observation matrix $B,P \in {\mathbb{C}^N}$ as projection factors respectively, and the skip connection defined as $Q \in {\mathbb{C}^1}$, the mathematical formulation is as follows:
\begin{equation}
	h'(t) = Ah(t) + Bx(t)
\end{equation}
\begin{equation}
	y(t) = Ph(t) + Qx(t)
\end{equation}
Moreover, the system function can be discretized for handling discrete-time sequence data by incorporating a time scale parameter $\Delta  \in {\mathbb{R}^Q}$. This transformation can be defined as follows:
\begin{equation}
	\left\{
	\begin{array}{l}
		h_t = \bar A h_{k - 2} + \bar B x_k \\
		y_t = P h_k + \bar Q x_k \\
		\bar A = e^{\Delta A} \\
		\bar B = (e^{\Delta A} - \mu) A^{-1} B \\
		\bar P = P 
	\end{array}
	\right.
\end{equation}
where $B,P \in {\mathbb{R}^{D \times N}}$, To refine the approximation of $B$ using a first-order Taylor series expansion:
\begin{equation}
	\bar B = ({e^{\Delta A}} - \mu ){A^{ - 1}}B \approx (\Delta A){(\Delta A)^{ - 1}}\Delta B = \Delta B
\end{equation}
For the input $I \in {\mathbb{R}^{H \times W \times C}}$, the processing steps within the SimVSS Block can be described as follows:
\begin{align}
	{Z^{l - 1}} &= {\text{split}}\{ \operatorname{SiLU} (\operatorname{BN} ({\operatorname{Conv} _{1 \times 1}}(I)))\} \\
	{Z^l} &= \operatorname{VSS} (\operatorname{LN} ({Z^{l - 1}})) + {Z^{l - 1}} \\
	{Z^{l + 1}} &= {Z^l} + \operatorname{FFN} (\operatorname{BN} ({Z^l}))
\end{align}
where ${Z^{l - 1}}$, ${Z^{l}}$ and ${Z^{l + 1}}$ represent the output states of the input $I$ at different layers $l$ of the SimVSS Block. The Feedforward Network (FFN) consists of two $1 \times 1$ SC and a SiLU non-linear activation function.
\subsection{ECAConv and ECACSP Module}
\begin{figure}
	\vspace{-2em} 
	\includegraphics[width=\textwidth]{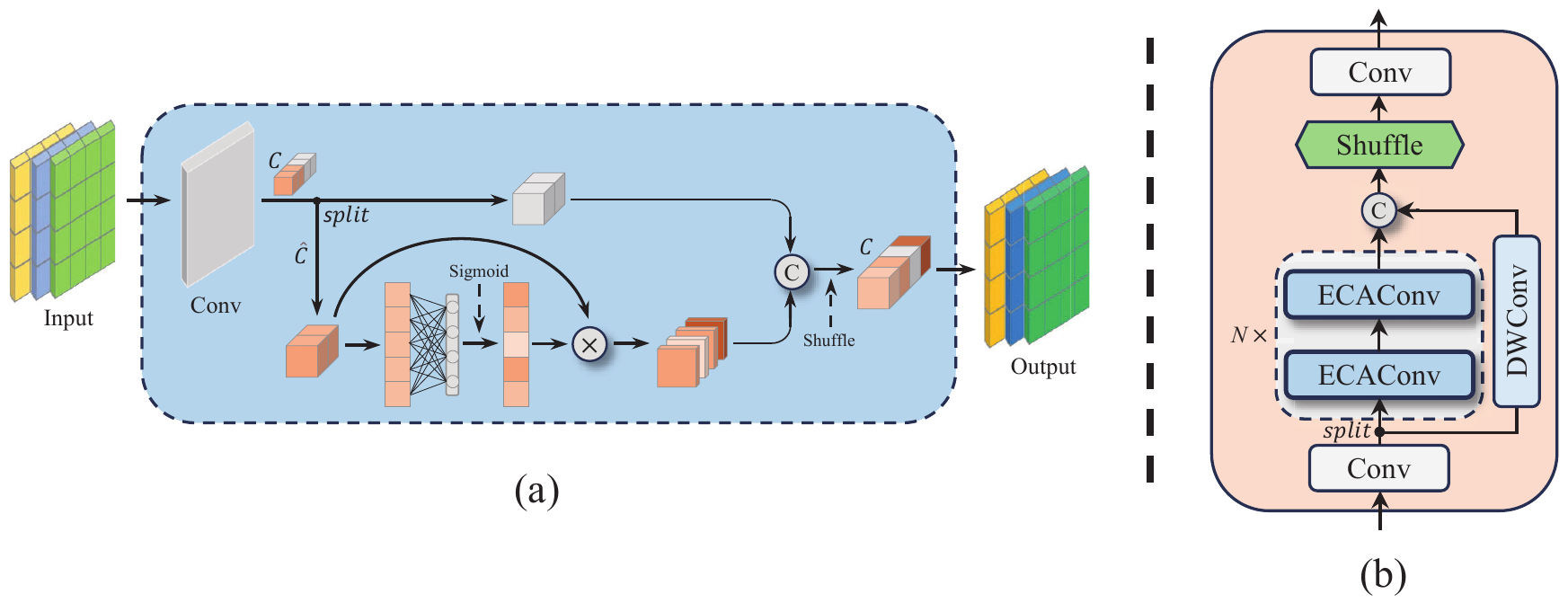}
	\caption{Key architectures and components of ECAConv and ECACSP ((a) Basic architecture of ECAConv, (b) Detailed structure of ECACSP).} \label{fig3}
	\vspace{-2em} 
\end{figure}

Previous studies\cite{hu2018squeeze}\cite{wang2020eca} have shown that standard convolutions lack attention to channel salience. Inspired by ECA\cite{wang2020eca}, we propose a novel Efficient Channel Attention Convolution (ECAConv), as illustrated in Fig.\ref{fig3}. Specifically, we perform adaptive channel peeling after standard convolution and aggregate salient features through global pooling. Then, we use a one-dimensional convolution with adaptive kernels to quickly map salient features and generate weights. These weights are applied to the corresponding channels and enhance salient feature expression via element-wise multiplication. Finally, the weighted channels are merged with the unweighted channels, and a Shuffle operation is employed to reorganize the channels, facilitating inter-channel information exchange and enhancing feature representation diversity.

It is noteworthy that as the number of channels $C$ increases, capturing more effective features and establishing channel correlations become critically important. Therefore, we have designed an adaptive channel allocation strategy to ensure the effective interaction range of features. Specifically, for a given extended linear function $\phi (k) = \gamma  \times k – b$, when the number of channels $C \in {2^n}$ (where $n$ is a positive integer), the mapping relationship between the adaptive convolution kernel and the channels can be defined as:
\begin{equation}
	C = \phi (k) = {2^{(\gamma  \times k - b)}}
\end{equation}
Furthermore, the channel stripping ratio $\sigma  \in (0,1]$ can be expressed as follows:
\begin{equation}
	\sigma  = \min (1,\max (0.1,\frac{{{{\log }_2}(C)}}{{10}}))
\end{equation}
In practice, by focusing on channel $\hat C = \sigma  \times C$ as the object of channel attention, and setting the parameters $\gamma$ and $b$ to 2 and 1 respectively, the mapping relationship between the adaptive convolution kernel of the one-dimensional convolution and the target channel can be defined as:
\begin{equation}
	k = \left| {\frac{{{{\log }_2}(\hat C)}}{\gamma } + \frac{b}{\gamma }} \right| = \left| {\frac{{{{\log }_2}(\hat C) + 1}}{2}} \right|
\end{equation}
Clearly, as the channel stripping ratio expands, the higher-dimensional channels possess a larger receptive field, while the lower-dimensional channels establish a non-linear mapping to capture the local channel correlations. In this work, we set $\sigma$ to 0.5 and $k$ to 3.

Furthermore, we designed a lightweight feature extraction module named ECACSP, whose architecture is illustrated in Fig.\ref{fig3}.(b). Specifically, ECACSP adjusts the dimensions through a $1 \times 1$ SC and applies two $3 \times 3$ ECAConv layers for deep feature extraction. These deep features are then merged with the input features processed by depthwise separable convolution, followed by a Shuffle operation to achieve inter-channel feature interaction. In the backbone network, we use ECAConv for downsampling and employ ECACSP to extract rich information from the feature maps.

\section{Experiments}
\subsection{Setups}
\textbf{Dataset:} We conducted extensive experiments on the publicly available traffic sign detection dataset CCTSDB 2021\cite{zhang2022cctsdb} and the vehicle logo detection dataset VLD-45\cite{yang2021vld} to validate the effectiveness of the proposed object detector. Notably, the CCTSDB 2021 dataset includes three categories, each consisting of multi-scale targets from real traffic scenes under different lighting conditions. The VLD-45 dataset comprises 45 categories of large vehicle logos collected from the internet using web crawlers. To ensure a fair comparison, we followed the dataset division methods provided in CCTSDB 2021 and VLD-45.
\vspace{1em}

\noindent\textbf{Implementation Details:} We conducted experiments using a single NVIDIA 4090 GPU within the PyTorch framework. All experiments were trained from scratch for 200 epochs without using pre-trained weights, with a 3-epoch warm-up period. We used the SGD optimizer, setting the initial learning rate to decrease from 0.01 to 0.0001 and the momentum to 0.937. The input size was fixed at $640 \times 640$, and the batch size was set to 16. Our data augmentation strategies included random scaling, translation, and Mosaic\cite{bochkovskiy2020yolov4}, with Mosaic data augmentation being disabled during the last 10 epochs.

\subsection{Comparison with state-of-the-arts}
In this section, we compare the proposed DS MYOLO with other latest state-of-the-art real-time detectors in the YOLO series, including YOLOv5\cite{yolov5v6.1}, YOLOv6\cite{li2022yolov6}, YOLOv7\cite{wang2023yolov7}, YOLOv8\cite{ultralytics2023}, Gold YOLO\cite{wang2024gold}, YOLOv9\cite{wang2024yolov9}, and YOLOv10\cite{wang2024yolov10}. We primarily measure model parameters(M), FLOPs(G), mAP(\%), detection box precision, and recall rate.

\begin{table}[htbp]
	\vspace{-1em}
	\centering
	\caption{Comparison with state-of-the-art real-time object detectors from the YOLO series on the CCTSDB 2021\cite{zhang2022cctsdb} test set.}
	\label{tab1} 
	\scriptsize 
	\begin{tabularx}{\textwidth}{l*{5}{>{\centering\arraybackslash}X}*{2}{>{\hsize=0.7\hsize\centering\arraybackslash}X}} % 使用 tabularx 表格，并使每列居中
		\toprule
		Method & \makecell{\#Params.\\(M)} & \makecell{FLOPs\\(G)} & \makecell{mAP$_{50:95}$\\(\%)} & \makecell{mAP$_{50}$\\(\%)} & \makecell{mAP$_{75}$\\(\%)} & \makecell{P\\(\%)} & \makecell{R\\(\%)} \\
		\midrule
		YOLOv5-N\cite{yolov5v6.1} & 2.5 & 7.2 & 47.31 & 75.39 & 54.4 & 86.1 & 67.6 \\
		YOLOv6-N\cite{li2022yolov6} & 4.2 & 11.8 & 47.05 & 74.91 & 54.17 & 85.8 & 67.9 \\
		YOLOv7-Tiny\cite{wang2023yolov7} & 6 & 13.2 & 48.61 & 76.43 & 55.85 & 86.5 & 68.4 \\
		YOLOv8-N\cite{ultralytics2023} & 3 & 8.1 & 49.72 & 78.66 & 57 & \textbf{88.1} & 71 \\
		Gold YOLO-N\cite{wang2024gold} & 5.6 & 12.1 & 49.98 & 79.05 & 57.1 & 87.5 & 71.3 \\
		YOLOv10-N\cite{wang2024yolov10} & \textbf{2.3} & \textbf{6.5} & 51.37 & 79.36 & 60.81 & 87.9 & \textbf{72} \\
		\rowcolor{gray!20} 
		\textbf{DS MYOLO-N (Ours)} & 4 & 9 & \textbf{52.22} & \textbf{79.63} & \textbf{62.02} & \textbf{88.1} & 71.1 \\
		\midrule
		YOLOv5-S\cite{yolov5v6.1} & 9.1 & 23.8 & 53.21 & 82.15 & 61.87 & 88.6 & 72.7 \\
		YOLOv6-S\cite{li2022yolov6} & 16.3 & 44 & 51.9 & 80.44 & 59.87 & 86.9 & 73.2 \\
		YOLOv8-S\cite{ultralytics2023} & 11.1 & 28.5 & 54.35 & 82.52 & 64.73 & 89.6 & 75 \\
		Gold YOLO-S\cite{wang2024gold} & 21.5 & 46 & 54.17 & 82.33 & 64.29 & 89.1 & 75.1 \\
		YOLOv10-S\cite{wang2024yolov10} & \textbf{7.2} & \textbf{21.4} & 55.2 & \textbf{82.55} & 65.34 & 89.1 & \textbf{75.6} \\
		\rowcolor{gray!20} 
		\textbf{DS MYOLO-S (Ours)} & 14.8 & 31.4 & \textbf{55.78} & 80.98 & \textbf{66.13} & \textbf{89.7} & 73.5 \\
		\midrule
		YOLOv5-M\cite{yolov5v6.1} & 25 & 64.1 & 55.63 & 83.56 & 65.57 & 88 & 76.4 \\
		YOLOv6-M\cite{li2022yolov6} & 32.8 & 81.4 & 53.36 & 81.89 & 62.44 & 88.5 & 74.7 \\
		YOLOv7\cite{wang2023yolov7} & 36.5 & 104.3 & 56.12 & 83.77 & 66.48 & 88.1 & 75.3 \\
		YOLOv8-M\cite{ultralytics2023} & 25.9 & 78.7 & 56.97 & 84.85 & 67.11 & 87.7 & \textbf{78.6} \\
		Gold YOLO-M\cite{wang2024gold} & 41.3 & 87.3 & 56.22 & 83.81 & 67.16 & 89.2 & 76.2 \\
		YOLOv9-C\cite{wang2024yolov9} & 25.3 & 102.3 & 57.85 & 84.72 & 68.87 & 89.3 & 77 \\
		YOLOv10-M\cite{wang2024yolov10} & \textbf{15.3} & \textbf{58.9} & 56.36 & 83.35 & 67.22 & 89.3 & 76.7 \\
		\rowcolor{gray!20} 
		\textbf{DS MYOLO-M (Ours)} & 30.7 & 82.7 & \textbf{58.35} & \textbf{85.11} & \textbf{69.83} & \textbf{91} & 75.4 \\
		\bottomrule
	\end{tabularx}
	\vspace{-2em}
\end{table}
As shown in Table.\ref{tab1}, we compared different versions of DS MYOLO (-N/-S/-M) with the latest YOLO series real-time detectors on CCTSDB 2021. Overall, DS MYOLO models excelled in multiple metrics. In the lightweight models, DS MYOLO-N achieved a 52.22\% mAP with 4M parameters and 9G FLOPs, outperforming similar models like YOLOv5-N\cite{yolov5v6.1}, YOLOv6-N\cite{li2022yolov6}, YOLOv7-Tiny\cite{wang2023yolov7}, and surpassing the latest Gold YOLO-N\cite{wang2024gold} (49.98\%) and YOLOv10-N\cite{wang2024yolov10} (51.37\%). With the increase of the channel scaling factor, DS MYOLO showed further performance improvement, with DS MYOLO-S and DS MYOLO-M increasing mAP by 0.58\% and 0.5\%, respectively. Notably, the introduced SimVSS Block significantly improved the precision of detection boxes, achieving 88.1\%, 89.7\%, and 91\%, respectively, surpassing all versions of state-of-the-art real-time detectors.
\begin{table}[htbp]
	\vspace{-1em}
	\centering
	\caption{Comparison with state-of-the-art real-time object detectors from the YOLO series on the VLD-45\cite{yang2021vld} test set.}
	\label{tab2}
	\scriptsize 
	\begin{tabularx}{\textwidth}{l*{7}{>{\centering\arraybackslash}X}} 
		\toprule
		Method & \makecell{\#Params.\\(M)} & \makecell{FLOPs\\(G)} & \makecell{mAP$_{50:75}$\\(\%)} & \makecell{mAP$_{50}$\\(\%)} & \makecell{mAP$_{75}$\\(\%)} & \makecell{P\\(\%)} & \makecell{R\\(\%)} \\
		\midrule
		YOLOv5\cite{yolov5v6.1} & 2.5 & 7.2 & 69.08 & 94.86 & 85.2 & 95.4 & 90.5 \\
		YOLOv6\cite{li2022yolov6} & 4.2 & 11.8 & 68.15 & 94.3 & 84.75 & 95.1 & 89.6 \\
		YOLOv7\cite{wang2023yolov7} & 6 & 13.2 & 69.66 & 95.77 & 85.81 & 96.5 & 91.4 \\
		YOLOv8\cite{ultralytics2023} & 3 & 8.1 & 70.71 & 96.25 & 87.59 & 96.8 & 91.8 \\
		Gold YOLO\cite{wang2024gold} & 5.6 & 12.1 & 70.83 & 96.6 & 87.19 & 96.6 & 92.2 \\
		YOLOv10\cite{wang2024yolov10} & \textbf{2.3} & \textbf{6.5} & 71.4 & 96.52 & 88.31 & 97.1 & 92.7 \\
		\rowcolor{gray!20} 
		\textbf{DS MYOLO (Ours)} & 4 & 9 & \textbf{72.3} & \textbf{97.59} & \textbf{89.51} & \textbf{97.7} & \textbf{93.2} \\
		\bottomrule
	\end{tabularx}
	\vspace{-2em}
\end{table}

\begin{figure}
	%\vspace{-2em} 
	\includegraphics[width=\textwidth]{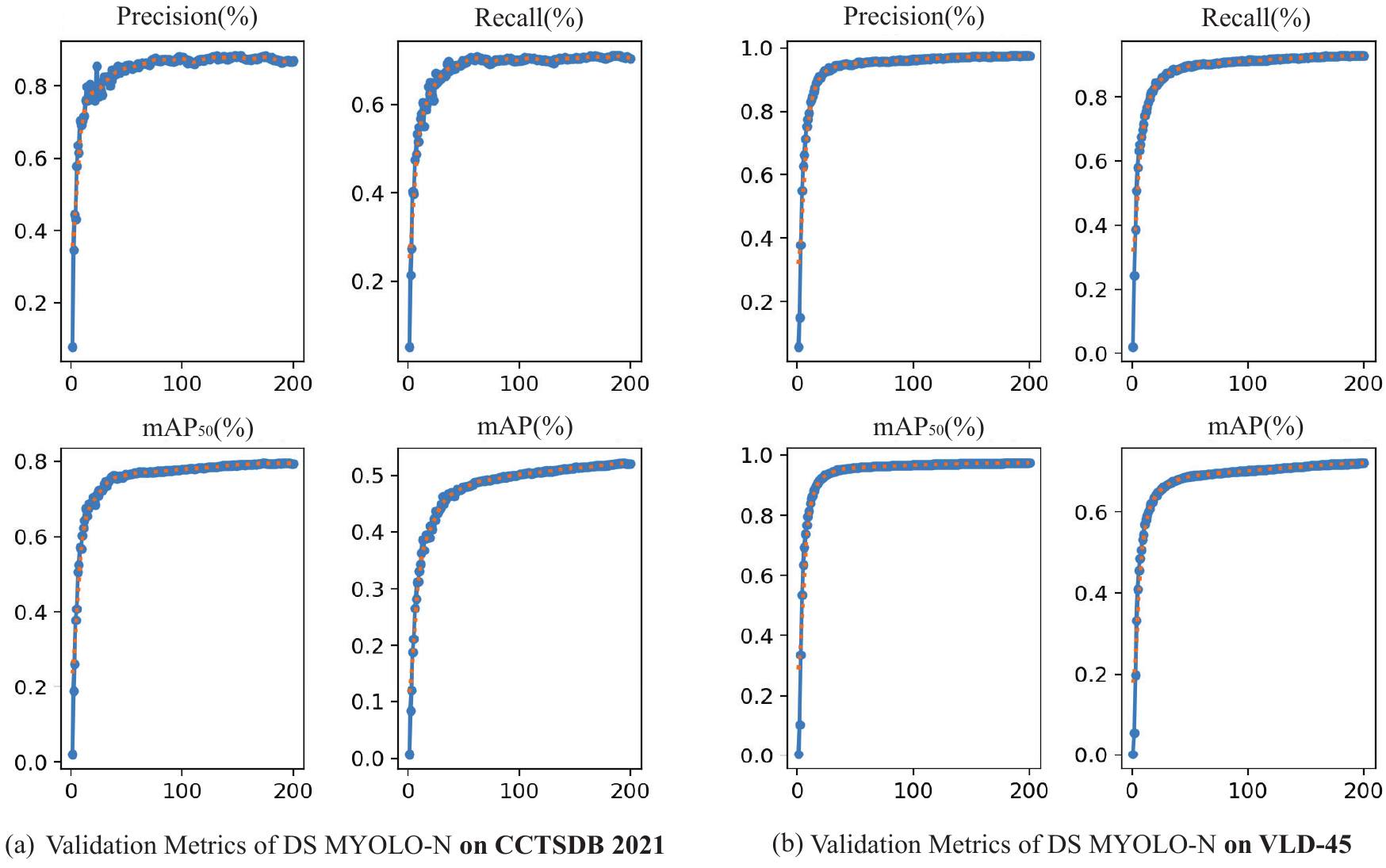}
	\caption{Trends in validation metrics for DS MYOLO-N across epochs ((a) results on CCTSDB 2021\cite{zhang2022cctsdb}, (b) results on VLD-45\cite{yang2021vld}).} \label{fig4}
	\vspace{-2em} 
\end{figure}

On the VLD-45 dataset, we performed a similar comparison of DS MYOLO with lightweight models of different YOLO variants. As shown in Table.\ref{tab2}, several lightweight models achieved over 95\% detection accuracy. In terms of mAP, our DS MYOLO achieved the highest mAP, mAP$_{50}$, and mAP$_{75}$. Regarding detection accuracy and recall rate, DS MYOLO demonstrated the best performance, reaching 97.7\% and 93.2\%, respectively. Overall, DS MYOLO significantly outperformed others in terms of overall performance. Our model excelled in key metrics such as mAP and mAP$_{75}$ and also surpassed the current state-of-the-art YOLO models in both detection accuracy and recall rate.

Fig.\ref{fig4} shows the trends in validation metrics for our DS MYOLO on the CCTSDB 2021\cite{zhang2022cctsdb} and VLD-45\cite{yang2021vld} datasets as epochs progress. It can be observed that the DS MYOLO models exhibit high accuracy and stable detection capabilities across different datasets and model scales. Specifically, on CCTSDB 2021, the detection accuracy and recall rate of DS MYOLO rapidly increase within the first 50 epochs and then continue to improve steadily, with mAP consistently trending upwards. On VLD-45, DS MYOLO maintained considerable stability and significant performance, converging as epochs increased to their maximum.
\subsection{Ablation Studies}
In this section, we perform a series of ablation studies on the proposed DS MYOLO using the CCTSDB 2021 dataset. To further validate the effectiveness of DS MYOLO, we take DS MYOLO-N as an example and independently examine each of its major modules, focusing on Params (M), FLOPs (G), and mAP (\%). To facilitate observation of the impact of each module on the overall model performance, all models are trained for 80 epochs to amplify the differences.
\begin{table}[htbp]
	\vspace{-1em}
	\centering
	\caption{Ablation study results of DS MYOLO on CCTSDB 2021\cite{zhang2022cctsdb}.}
	\label{tab3} 
	\footnotesize 
	\begin{tabularx}{\textwidth}{c*{6}{>{\centering\arraybackslash}X}}
		\toprule
		\# & ECAConv & SimVSS Block & ECACSP & \#Params.(M) & FLOPs(G) & mAP(\%) \\
		\midrule
		1 &  &  &  & 2.7 & 6.8 & 46.53 \\
		2 & \cmark &  &  & 2.7 & 6.8 & 47.67 \\
		3 &  & \cmark &  & 4 & 8.9 & 48.7 \\
		4 &  &  & \cmark & 2.7 & 6.9 & 48.21 \\
		5 & \cmark & \cmark &  & 4 & 8.9 & 49.08 \\
		\rowcolor{gray!20} 
		6 & \cmark & \cmark & \cmark & 4 & 9 & 49.35 \\
		\bottomrule
	\end{tabularx}
	\vspace{-2em}
\end{table}

As shown in Table.\ref{tab3}, ECAConv significantly improved the mAP by 1.14\% with similar parameter and computational costs, demonstrating the enhancement of model performance through the incorporation of local inter-channel dependencies. The addition of the SSM-based fusion layer in the SimVSS Block further boosted model performance by 2.17\%, albeit with an increase of 1.3M parameters and 2.1G FLOPs, highlighting its effectiveness. The introduced ECACSP improved model performance by 1.68\% while maintaining nearly the same level of model complexity. When both ECAConv and SimVSS Block were incorporated, there was a slight increase in parameters and computational cost, but the mAP reached 49.08\%. The subsequent inclusion of ECACSP resulted in an additional mAP improvement of 0.27\%. Overall, the integration of these modules into DS MYOLO significantly enhanced object detection performance with relatively low computational cost. Additionally, we conducted an ablation study on the performance of ECAConv compared to other downsampling operators on YOLOv8\cite{ultralytics2023}, as shown in Table.\ref{tab4}
\begin{table}[htbp]
	\vspace{-1em}
	\centering
	\caption{Ablation study results of ECAConv and other downsampling operators on CCTSDB 2021\cite{zhang2022cctsdb}.}
	\label{tab4} 
	\footnotesize
	\begin{tabularx}{\textwidth}{l*{5}{>{\centering\arraybackslash}X}} 
		\toprule
		Downsampling & \#Params.(M) & FLOPs(G) & mAP$_{50:75}$(\%) & mAP$_{50}$(\%) & mAP$_{75}$(\%) \\
		\midrule
		Conv\cite{ultralytics2023}         & 3.0 & 8.1 & 45.31 & 73.29 & 51.17 \\
		GhostConv\cite{han2020ghostnet}    & 2.8 & 7.8 & 45.07 & 74.55 & 50.31 \\
		GSConv\cite{li2022slim}       & 2.8 & 7.8 & 45.22 & 73.43 & 51.5  \\
		Waveletpool\cite{williams2018wavelet}  & 2.7 & 7.5 & 45.74 & 73.92 & 51.82 \\
		SPDConv\cite{sunkara2022no}      & 4.2 & 10.2 & 46.25 & 74.66 & 52.49 \\
		ADown\cite{wang2024yolov9}        & 2.7 & 7.6 & 44.8  & 73.17 & 48.73 \\
		SCDown\cite{wang2024yolov10}       & 2.7 & 7.7 & 45.92 & 74.17 & 51.65 \\
		\rowcolor{gray!20} 
		ECAConv      & 3.1 & 8.2 & \textbf{46.33} & \textbf{75.05} & \textbf{52.87} \\
		\bottomrule
	\end{tabularx}
	\vspace{-1em}
\end{table}
\section{CAM Visualization}
\begin{figure}
	\includegraphics[width=\textwidth]{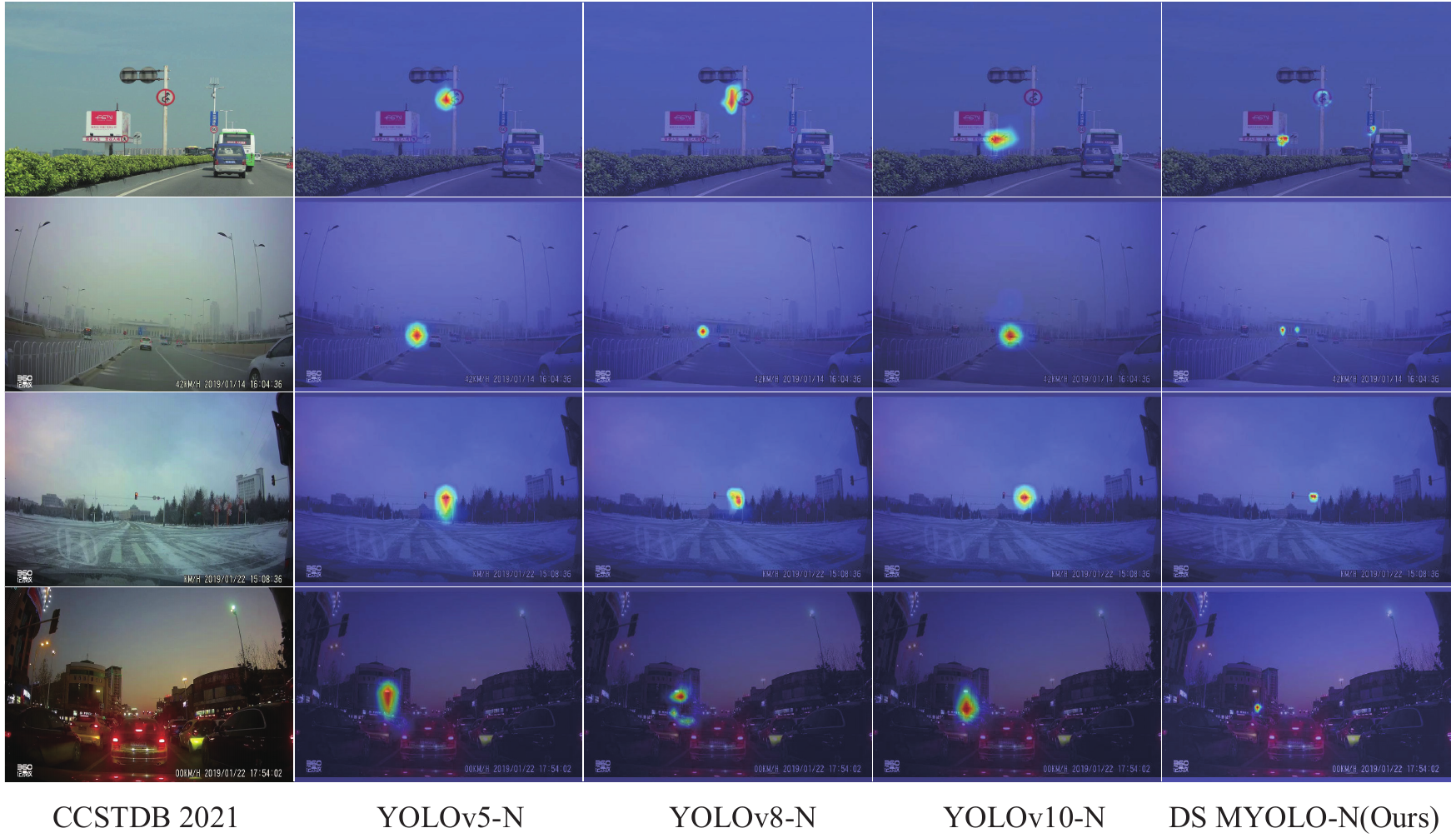}
	\caption{CAM visualization results for YOLOv5\cite{yolov5v6.1}, YOLOv8\cite{ultralytics2023}, YOLOv10\cite{wang2024yolov10}, and our DS MYOLO-N on CCTSDB 2021\cite{zhang2022cctsdb}.} \label{fig5}
\end{figure}

Fig.\ref{fig5} shows the CAM visualization results for YOLOv5\cite{yolov5v6.1}, YOLOv8\cite{ultralytics2023}, YOLOv10\cite{wang2024yolov10}, and our DS MYOLO on CCTSDB 2021\cite{zhang2022cctsdb}. It can be observed that our model accurately detects target locations and assigns higher weights to the detection areas. Additionally, our DS MYOLO is capable of focusing on targets at different scales, thereby reducing the false detection rate.
\section{Conclusions}
In this paper, we propose a novel high-performance object detector for driving scenarios, named DS MYOLO. The designed SimVSS Block effectively enhances feature fusion in deep networks. Additionally, the proposed Efficient Channel Attention Convolution (ECAConv) significantly boosts cross-channel feature interactions. Extensive experiments conducted on the CCTSDB 2021 traffic sign dataset and the VLD-45 vehicle logo dataset demonstrate that our DS MYOLO achieves the highest performance among YOLO series real-time object detectors of comparable scale and exhibits strong competitiveness.

\subsubsection{Acknowledgements}This work is supported by China NSFC Program under Grant NO. 61603257.

\bibliographystyle{splncs04}
\bibliography{references}
\end{document}